# Semantic Search of Memes on Twitter

Jesus Perez-Martin [1,2], Benjamin Bustos [1,2], Magdalena Saldana [1,3]

[1]Millennium Institute of Data Foundation, Santiago, Chile
[2]Departament of Computer Science, University of Chile
[3]Communication Faculty, Pontifical Catholic University of Chile

***Abstract***

Memes are becoming a useful source of data for analyzing behavior on social media. However, a problem to tackle is how to correctly identify a meme. As the number of memes published every day on social media is huge, there is a need for automatic methods for classifying and searching in large meme datasets. This paper proposes and compares several methods for automatically classifying images as memes. Also, we propose a method that allows us to implement a system for retrieving memes from a dataset using a textual query. We experimentally evaluate the methods using a large dataset of memes collected from Twitter users in Chile, which was annotated by a group of experts. Though some of the evaluated methods are effective, there is still room for improvement.

## 1. Introduction

Memes are cultural expressions that are transmitted using visual material. In the fields of Computer Vision and Multimedia Information Retrieval, Internet memes have attracted the attention of the research community (Bauckhage, 2011; Bordogna & Pasi, 2012; Milo et al., 2019; Chew & Eysenbach, 2010; Ferrara et al., 2013; JafariAsbagh, 2014; Finkelstein et al., 2018; Peirson & Tolunay, 2018; Truong et al., 2012; Tsur & Rappoport, 2015; Zannettou et al, 2018). And their adequate classification and retrieval, as opposed to general images, are challenging problems since they have a more complex semantic meaning that is often ironic and ambiguous.

This study observes meme usage on Twitter in Chile, a country with high rates of internet penetration and social media usage (We Are Social, 2019). Our long-term goal is to develop algorithms that allow us to determine which images are based on memes (or all the memes that have been created with the same image), for observing the discourse associated with said memes. For example, the so-called Drake's meme (see Fig. 1) compares two images to emphasize correct actions versus incorrect or undesirable actions: study in advance versus study the night before, use Photoshop versus non-professional software, etc.

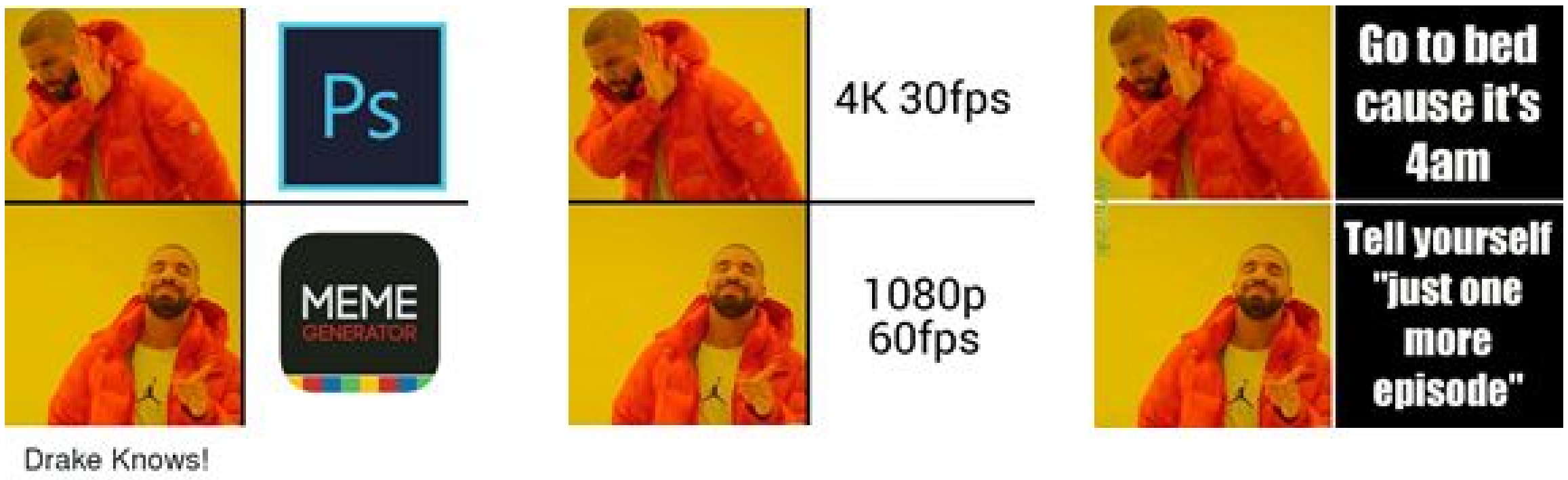

**Figure 1.** An example of the Drake's meme

This meme has been adapted (see Fig. 2) to the local reality using an image inspired by Drake and texts related to the Chilean context ("abuela"=grandmother, "awela"=a mispronunciation of the word).

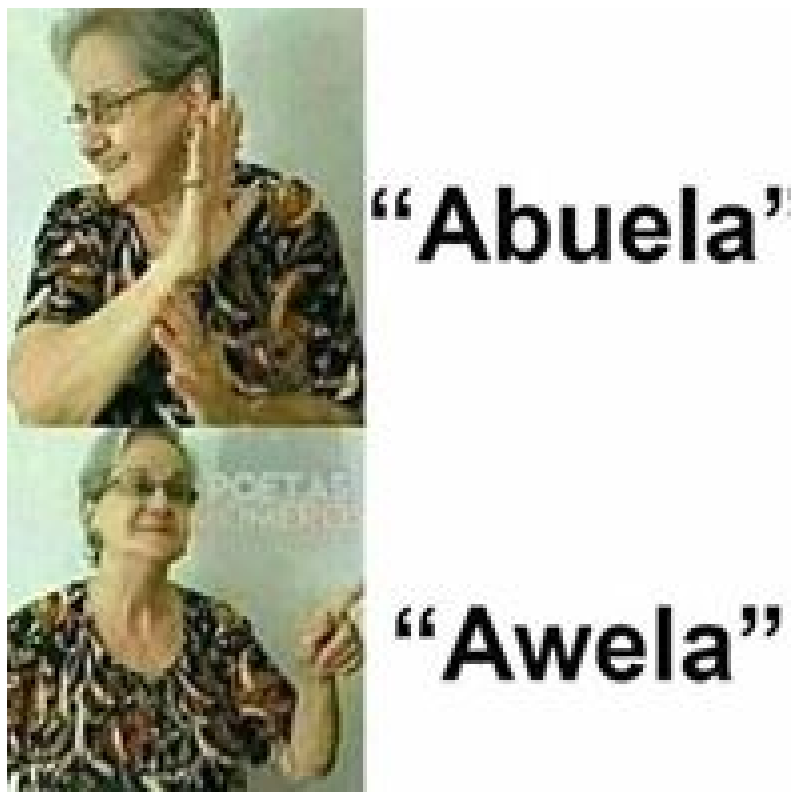

**Figure 2.** Local adaptation (Chilean context) of the Drake's meme

But before observing how memes are adopted and modified by local contexts, we first need to be able to automatically classify images as memes. By using machine learning techniques, this study offers a step-by-step process to explain how to collect memes containing images and how to classify images according to whether these are memes or not. First, we collect a large dataset of Chilean tweets with images, where we manually annotate an important portion of those images. Second, we develop a robust classification algorithm to classify images on the Internet and detect memes and stickers. Third, we propose a deep learning model capable of retrieving memes from a natural language description in Spanish and vice-versa. Finally, we present an experimental evaluation of our proposed algorithms. Additionally, we will release our processing algorithms and annotated meme dataset to the research community.

### 1.1. Definition of Meme

For Gleick (2011), memes are nothing more than ideas, images, slogans, melodies, stories, recipes, skills or abilities, legends, and systems that populate our minds, which use our brains as computers that give them life. We are their vehicles and those that make them possible. They are transmissible from "word of mouth"; on clay tablets, cave walls, sheets of paper; using pencils and pens and in our printers; on magnetic tapes, optical discs, and broadcasting towers.

Nowadays, a meme has been defined as an image, video, or text that uses humor as a resource to communicate an idea or message, which is disseminated through the Internet mainly via social media (Davison, 2012). Based on Chagas, Freire, Ríos, & Magalhães (2019) and García (2014), we consider a meme as an image with the following characteristics:

- It consists of an image (photo, illustration, scheme, etc.) with an overlay text. If any of these elements is missing, it will not be considered to be a meme.

- The superimposed text must be a message or motto that is humorous. This element of humor can be a parody, irony, or a joke.

- It presents a high level of intertextuality. This means that the image or text refers to an event, icon, or phenomenon of current affairs (news) or popular culture (e.g., movies, series, commercials, other memes, etc.).

- It should be taken into account that memes are cultural phenomena, so they respond to the social reality of the people who create and share them. This means that there may be cases in which it is necessary to perform an interpretation to decide whether it is a meme or not.

- Some memes mimic WhatsApp stickers, a feature introduced recently by WhatsApp for users to communicate using visuals beyond the popular "emoji." Stickers are images that may or may not have text. In the case of memes, sticker-style memes are images with a text working as a short answer, such as "yes", "thank you" or "love ya." This type of meme does not necessarily carry humor or intertextuality. As such, we decided to treat it as a different meme category.

## 2. Basic Concepts

### 2.1. Tasks

To be able to capture images posted on Twitter and automatically classify them as memes, we defined some tasks that we consider relevant for this purpose:

- To generate a system that identifies Chilean accounts and download all tweets that have images.

- To generate an algorithm capable of automatically classifying images as a *meme*, *sticker*, and *no-meme* classes.

- To generate an algorithm able to rank and match a set of memes given a sentence as a query.

To accomplish these tasks, we need to use different kinds of machine learning algorithms. These algorithms help us to discover patterns in the data and classify the images into two groups (*meme* / *no-meme*). There are two main types of techniques, based on the way they “learn” about data to make predictions: a) supervised techniques, and 2) unsupervised techniques.

### 2.1.1. Classification

The classification task uses supervised learning and it is the kind of algorithm we need to accomplish the second task (generate an algorithm for automatically classifying images as *meme* / *no-meme*). Supervised learning is where we have input variables $X$ and an output variable $Y$ and we want to learn the mapping function from the input to the output $Y = f(X)$. The goal is to approximate the mapping function, so that when one has new input data $X'$ one can predict the output variables $Y'$ for that data. In other words, in a classification task, we are given a *pattern* and the task is to classify it into one out of $c$ classes (Theodoridis, Pikrakis, Koutroumbas, & Cavouras, 2010). The number of classes, $c$, is assumed to be known a priori.

### 2.1.2. Semantic Search

To accomplish the third task, we need to develop a *search with meaning*. This search is different from *lexical search* where the search engine looks for literal matches of the query words or variants of them, without understanding the overall meaning of the query. In the case of memes, we need to improve the search accuracy by understanding the searcher's intent and the contextual meaning of terms as they appear in the searchable dataspace, either on the Web or within a closed system, to generate more relevant results. For this, we need to consider various points including the description of memes, their semantic interpretation, the text into them, the variation of words, synonyms, generalized and specialized queries, concept matching, and natural language queries to provide relevant search results.

## 2.2. Vector Model

To represent the properties or characteristics (patterns) of a phenomenon being observed, a common approach is using a set of feature values $x_i$, $i = 1, 2, ..., l$, which make up the $l$ -dimensional feature vector $x = [x_1, x_2, ..., x_l] \in R^l$. This general approach is known as the vector model, and it is widely used for characterizing multimedia data (images, text, etc.). Feature vectors are also known as “descriptors”. In many cases, this numerical representation facilitates processing the data and performing statistical analysis [1]. For representing images as vectors, the feature values might correspond to the data in the pixels of the image. On the other hand, for representing text the feature values might correspond to the frequencies of occurrence of textual terms in a document.

## 2.3. Visual Descriptors

Next, we describe the visual descriptors that we use in this study.

### 2.3.1. Histogram of Oriented Gradients

Dalal and Triggs (2005) proposed the so-called Histogram of Oriented Gradient (HOG) descriptor. It has been widely and successfully used in tasks like object detection, where it was the state of the art before the deep learning era. HOG represents an image as a single feature vector.

The basic idea of HOG is that local object appearance and shape can often be characterized rather well by the distribution of local intensity gradients or edge directions, even without precise knowledge of the corresponding gradient or edge positions [2]. HOG works with something called a block, which is similar to a sliding window. A block is a pixel grid in which gradients are constituted from the magnitude and direction of change in the intensities of the pixel within the block.

### 2.3.2. ResNet

With the deep learning era and the Convolutional Neural Networks, researchers have presented several visual descriptors that are the state of the art in computer vision tasks. K. He et al. (2016) proposed the Residual Neural Network (ResNet) that utilizes *skip connections* or *short-cuts* to jump over some layers of the network to extract low, middle and high-level features. These networks have achieved the human level image classification result.

## 2.4. Text Embedding

Besides the image representation, we need to use mathematical representations for text (text embedding). This is required to implement the semantic search of memes from natural language queries. The text embeddings are created by analyzing a set of texts and representing each word, phrase, or entire document as a vector in a high dimensional space.

A simple way of representing a text is by computing the average of the vector representation of each word (word embedding). A word embedding is a learned representation for text where words that have the same meaning have a similar representation. Researchers have proposed several useful word embeddings and methods like GloVe (Pennington et al, 2014), Word2Vec (Mikolovet al., 2013) and FastText (Bojanowski et al., 2017). Specifically, for this work, we need to represent text in Spanish. For this, the repository[1] includes links to useful Spanish word embeddings computed with the methods mentioned before.

## 3. SemanticMemes Dataset

An important contribution of this work is the construction of a large dataset, that we called *SemantcMemes*[2], of textually described memes in Spanish. To build this dataset, we started collecting all Chilean tweets with images posted from May 10th, 2019 until now. Then, four experts manually classified a subset of 52,000 of a total of more than 518,284 images. For this purpose, the experts were trained until they reached inter-coder reliability (ICR) of 90% or above, and a Krippendorff's alpha of .7 or above. ICR was calculated on a subsample of 2,000 images not included in the final sample of 52,000 cases. Each expert classified 13,000 different images into four classes: *Meme*, *No-meme*, *Sticker*, and *Doubtful*. Also, for images classified as *Memes* and *Stickers*, the experts transcribed the text in the image, provided a description of the contents in the image (subjects, places, colors, actions, etc.), and a semantic interpretation of the context and situation showed.

[1] Link to the implementation of our models and training methods: https://github.com/jssprz/SemanticMemes.git

[2] Website of our SemanticMemes dataset: https://jssprz.github.io/semantic-memes-docs/dataset/

Currently, our generated dataset contains 52,000 images: 1,194 memes, 1,443 stickers, 49,347 no-meme, and 16 doubtful. The transcriptions, descriptions, and annotations from the images conform to a vocabulary of 8,669 words.

## 4. Meme Recognition Task

Being able to predict the kind of image that is used in a blog post or a tweet could be useful for analyzing the semantics of the whole content. As such, one of the main contributions of this paper is the creation of a classification algorithm capable of classifying any image among the classes *meme*, *sticker* and *no-meme*.

To classify the images into the previously mentioned classes several techniques could be used, like background segmentation or text detection. In this work, we propose to extract a global visual feature descriptor of the image like HOG or ResNet and apply a machine learning classifier like Decision Tree (Quinlan, 1986), KNN (Cover & Hart, 1967), SVM (Schölkopf, 1998) or Neural Networks (Goodfellow et al., 2016). We obtained the best results using the ResNet features and the SVM classifier.

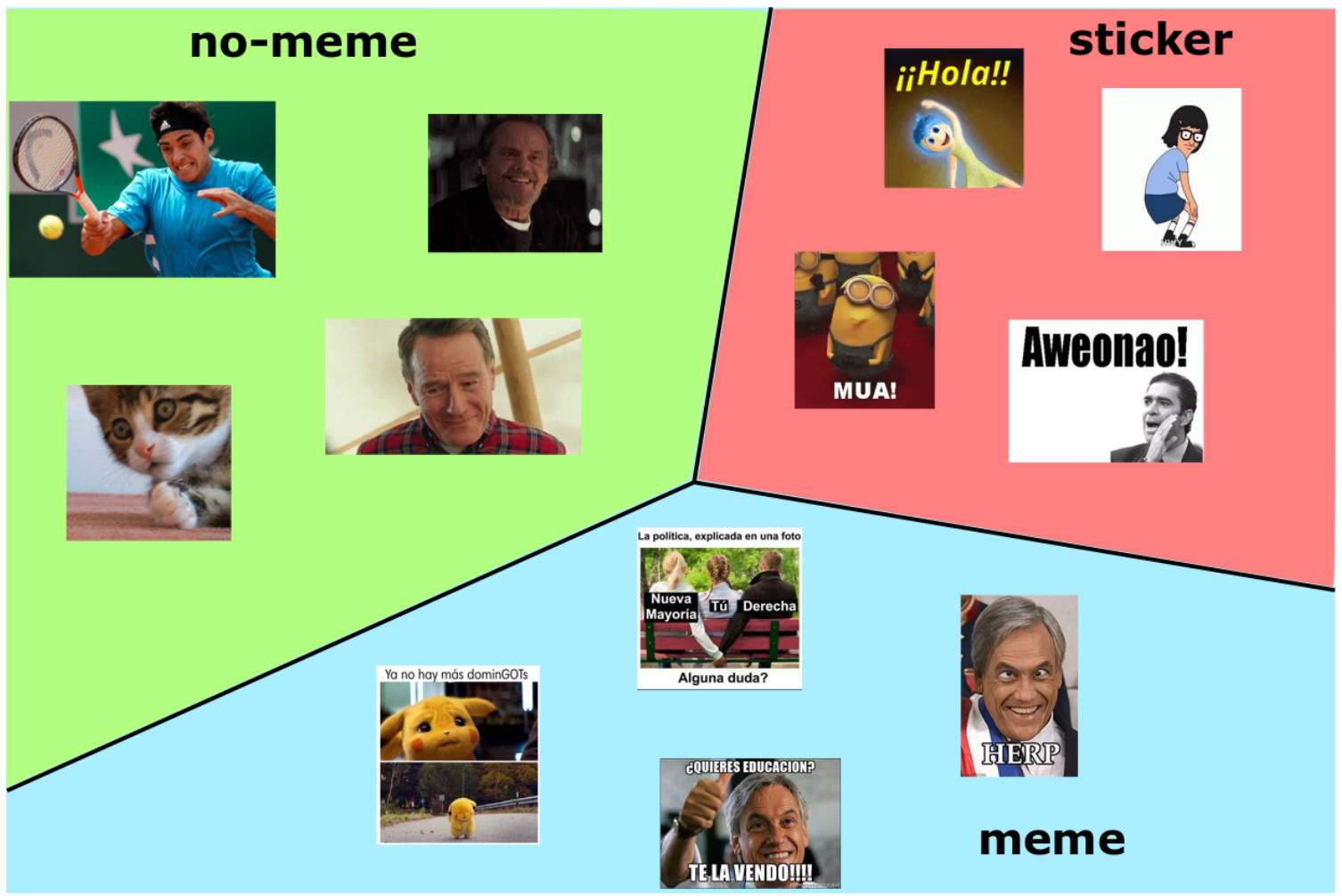

**Figure 3.** We propose to learn separating hyperplanes by SVM machine learning classifier between the *no-meme*, *sticker* and *meme* images.

## 5. Meme Retrieval Task

The search for memes, or "meme retrieval task", is increasing the attention of researchers (Milo et al., 2019). As opposed to general image retrieval, a user searching for a meme may have multiple search criteria like the background image, the text in the meme, the description, and/or the interpretation. The relevance of each criterion may vary. In our experiments, we show that a general image retrieval algorithm cannot be used effectively for the meme retrieval task.

We now introduce our semantic search of memes model, see Fig. 3. The idea is to build a common visual-semantic-embedding (Fang et al., 2015, Miech et al., 2018). To build this embedding, the model combines a *language model* to map the text captions to a language representation vector (text embedding) and a *visual model* to obtain a visual representation vector. Then, the model projects the visual representation vector and language representation vector into a shared feature space, minimizing the distance between them. During the inference time, the model maps the input sentence description to a point in the shared space corresponding to a semantically close meme and vice-versa.

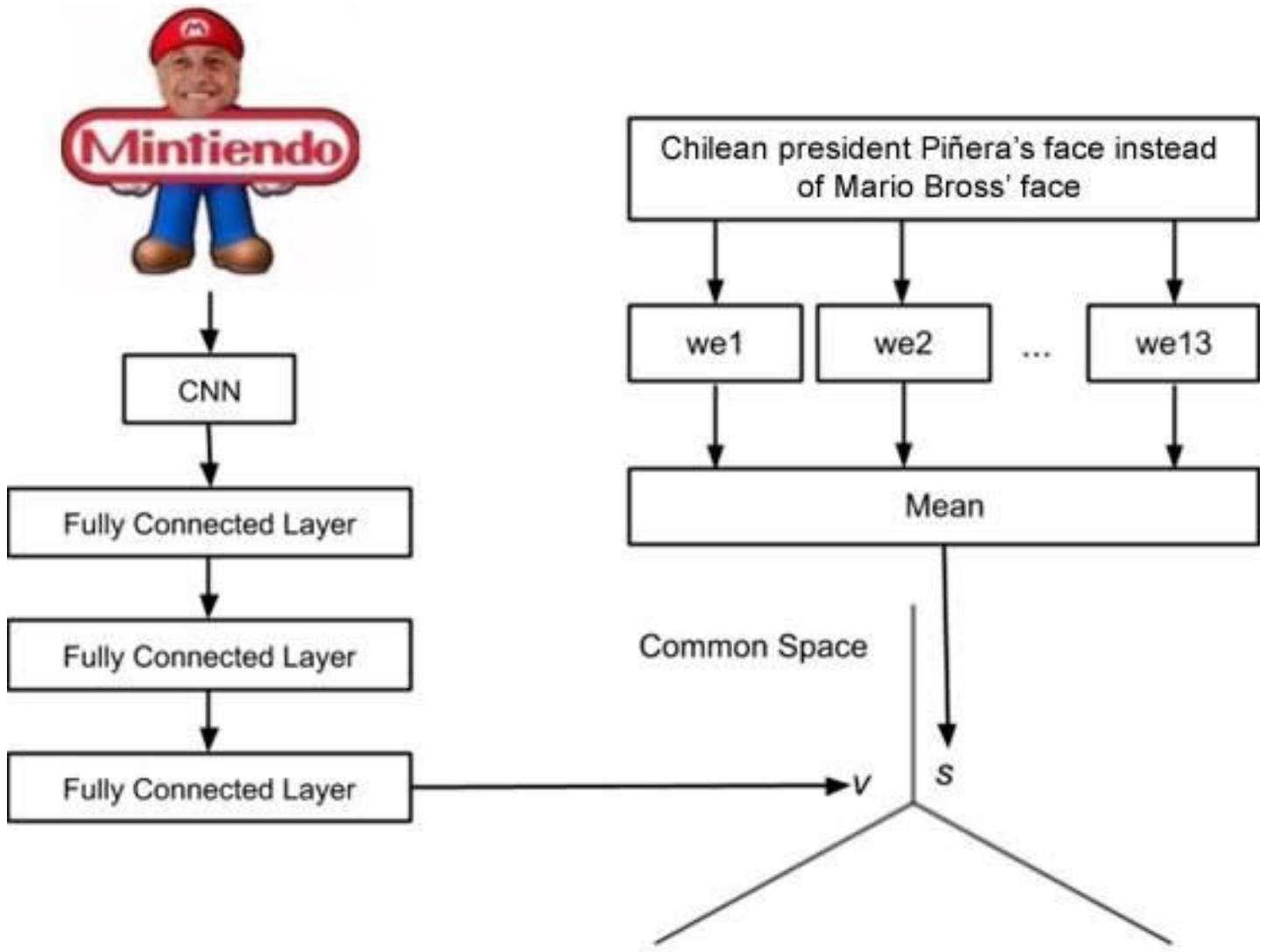

**Figure 3.** The meme retrieval model

## 5.1. Visual Model

On the one hand, to process the visual content of images (the image branch in Fig. 3), we propose to use a deep convolutional network (CNN). Specifically, we use ResNet [3] with 152 layers. This network and its variants have achieved great success in a wide range of vision problems (Yan & Mikolajczyk).

## 5.2. Language Model

On the other hand, to represent the Spanish texts in the shared space (the text branch in Fig. 3), we propose to use the FastText model (Bojanowski et al., 2017) pre-trained on the Spanish Billion Word Corpus[3] with 1.4 billion words. This model provides a 300-dimensional vector representation for each word. We represent the whole text as the average of these vectors.

[3] http://crscardellino.github.io/SBWCE/

### 5.3. Loss Function

In the context of deep learning and optimization algorithms, the function used to evaluate a candidate solution is referred to as the **objective function** or **criterion**. We want to maximize or minimize this function. When we are minimizing it, we may also call it the **cost function**, **loss function**, or **error function** (Goodfellow et al., 2016).

To obtain a model capable to rank the results according to the similarity of the memes with the query text, the model must learn to distinguish between similar and dissimilar objects. To do this, a straightforward idea is feeding the system with one positive and one negative example at a time, adding up the losses:

$$L = L_{-} + L_{+}$$

A better idea is to compare a baseline sample of the dataset ($a$) with a random positive sample ($p$) and a negative sample ($n$), minimizing the distance to the positive $d(a, p)$ and maximizing the distance to the negative $d(a, n)$. This alternative method is known as **triplet loss** (Schroff et al., 2015), and it is formulated as:

$$L = max(d(a, p) - d(a, n) + m, 0),$$

where $d$ is a distance function (e.g., the Euclidean distance ($L^2$) and $m$ is an arbitrary margin used to further the separation between the positive and negative scores.

## 6. Experimental Evaluation

### 6.1. Meme Recognition Experiments

Firstly, to evaluate our meme Recognition approach we need to note that our dataset presents an unequal distribution of classes, with a ratio of 50:1 between classes *no-meme* and *meme*. If we train a classification model without fixing this problem, the model will be completely biased. To solve this problem, we could use some techniques like **Resampling** (**Oversampling** or **Undersampling**). In our experiments, we use Undersampling.

The idea is to randomly delete some samples of the majority class (no-meme) until matching with the other classes (meme and sticker). Fig. 4 shows the Dataset distribution before and after undersample.

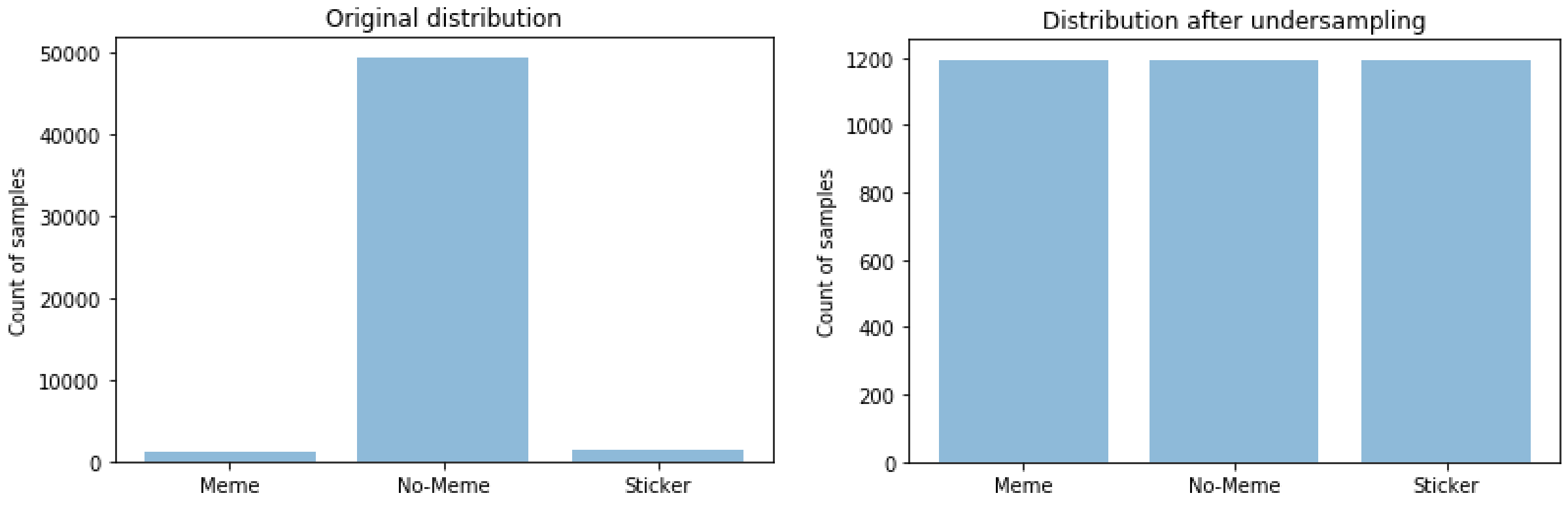

**Figure 4.** Dataset distribution

To improve the quality of our results, we evaluate each classification method repeating the random undersampling process 10 times and running a stratified 10-fold cross-validation for each resample. The 10-fold cross-validation consists of randomly partitioning the resample into 10 equal-sized subsamples. A single subsample is retained as the validation data for testing the model, and the remaining 9 subsamples are used as training data. The cross-validation process is then repeated 10 times, with each of the 10 subsamples used exactly once as the validation data. Finally, we compute the average among the 10 results to produce a single estimation.

We measure the performance of the classifiers attending to:

**Precision:** It is the ratio $tp/(tp+fp)$ where $tp$ is the number of true positives and $fp$ the number of false positives. The precision is intuitively the ability of the classifier not to label as positive a sample that is negative. The best value is 1 and the worst value is 0.

**Recall:** It is the ratio $tp/(tp+fn)$ where $fn$ is the number of false negatives. The recall is intuitively the ability of the classifier to find all the positive samples. The best value is 1 and the worst value is 0.

**F1-score:** It is the weighted average of the precision and recall, where an $F1$ score reaches its best value at 1 and the worst score at 0. The relative contribution of precision and recall to the $F1$ score are equal. The formula for the $F1$ score is $F1 = 2 * \frac{precision * recall}{precision + recall}$

The **Confusion Matrix** $C$ is such that $C_{i,j}$ is equal to the number of observations known to be in group $i$ but predicted to be in group $j$.

Tab. 1 shows that the ResNet + LinearSVM method obtains the best results with a precision of 0.73. Also, Fig. 5 shows that the model is capable of classifying the sticker class correctly in 75% of cases, but the model wrongly classifies the 15% of meme samples as sticker.

| Method | Precision | Recall | F1-Score |
|---|---|---|---|
| HOG + Decision Tree | 0.533 | 0.534 | 0.533 |
| HOG + Naive Bayes | 0.556 | 0.554 | 0.547 |
| HOG + KNN | 0.554 | 0.55 | 0.546 |
| HOG + Linear-SVM | 0.574 | 0.574 | 0.57 |
| ResNet + Decision Tree | 0.651 | 0.649 | 0.649 |
| ResNet + Naive Bayes | 0.641 | 0.633 | 0.631 |
| ResNet + KNN | 0.7 | 0.676 | 0.672 |
| **ResNet + LinearSVM** | **0.73** | **0.73** | **0.73** |

**Table 1.** Average results of *Meme* / *Sticker* / *No-Meme* classification on the Twitter Memes Dataset

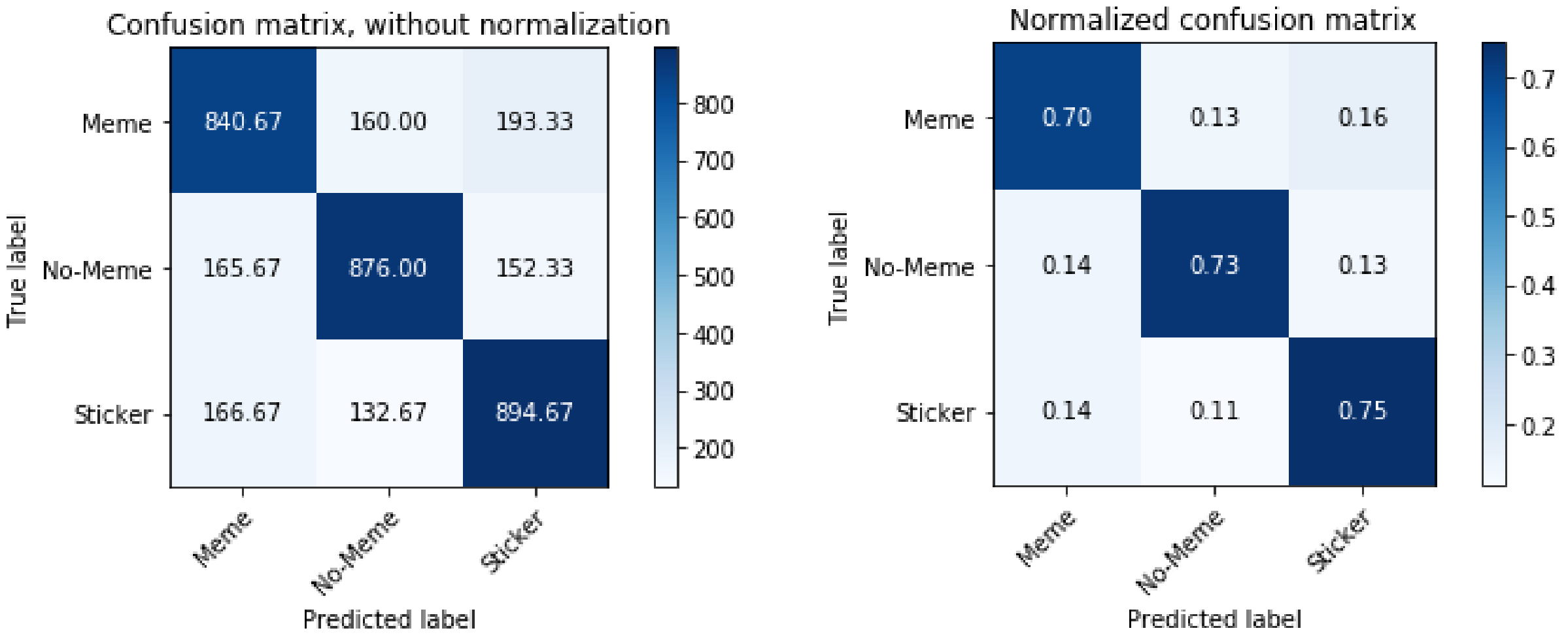

**Figure 5.** Average Confusion Matrix results of ResNet + KNN method

### 6.2. Meme Retrieval Experiments

In this section, we evaluate the proposed model to deal with the Meme Retrieval From Text task. To do this is required to split the dataset into training and testing subsets. We divide the dataset according to the(meme, text) pairs into a set of 6,228 pairs for training and 83 pairs for testing.

We trained the model for 270 epochs with a batch size of 16. We used the Stochastic Gradient Descent algorithm as optimizer with a learning rate of 0.0001. We used Triplet Loss as the loss function with a margin parameter of 1.0. We measure the performance of the model using the **mean average precision** (mAP) score. This score summarizes a precision-recall curve as the weighted mean of all precision values achieved at each threshold, with the increase in recall from the previous threshold as the weight. In other words, a higher value of mAP implies the model is better ranking the results and putting the most similar objects in the first positions.

Fig. 6 shows how the model reaches the highest mAP on the test set of 0.3 after 270 epochs. Analyzing the curve of loss, we observe that the model never overfit on the training set. Thus, we could achieve better results if we continue the training process for more epochs.

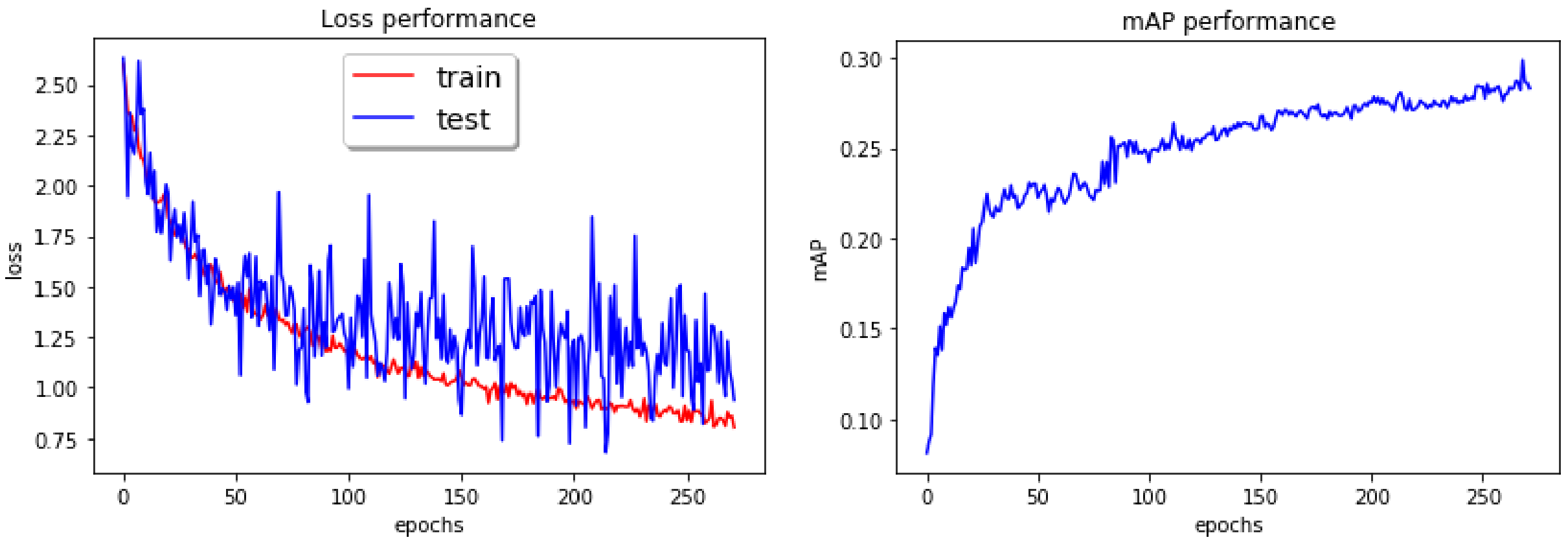

**Figure 6.** Training process. The loss curves on train and test sets, and mAP curve on test set.

## 7. Conclusions and Implications

We propose and compare methods for automatically classifying images as *meme*, *no-meme*, or *sticker*. Our experimental evaluation, using a large dataset of memes obtained from Chilean Twitter and manually annotated by experts, shows that deep learning methods achieve higher effectiveness than classical methods for feature representation. Nevertheless, the best-tested method achieves 73% of precision, which is far from perfect. Thus, there is room for continuing researching and developing better automatic classification methods.

Also, we propose a method for performing a semantic retrieval of memes using textual queries. For achieving this, we trained a neural network for describing images from memes and sentences (text) in a common feature space. This allows us to produce a ranking of memes in our dataset according to their relevance to the textual query. Our experiments show that the model is capable of learning the similarities among texts and memes. By training for more epochs one could achieve better results. However, there are other techniques like Recurrent Neural Networks (RNN) and Long Short-Term Memory (LSTM) Networks (Hochreiter & Schmidhuber) that could be evaluated for the text embedding branch.

Different parts of the memes (caption, image or user tags) can be used to define different search criteria (Milo et al., 2019) and obtain better results. In our dataset, we have access to information about the tweets where the images were posted (text of tweets, tags, location, user, etc.). However, in this paper, we only use the content of the image for all the tasks. In our future work, we want to evaluate the use of context information for the meme retrieval task.

This paper has made at least to contributions to the field of computer methods in communication. First, it provides a method to automatically retrieve and classify images from Twitter into *memes*, *stickers*, and *no-memes*. Nowadays, the use of memes on social media have become increasingly prevalent, but the volume of information and the velocity data are generated makes it difficult for researchers to be able to collect and analyze such amounts of information, especially if the analysis is conducted manually. This study offers a path for automatically code images in reliable ways. Second, we offer access to the dataset we retrieved and analyzed for researchers to use it. The dataset will be available after the blind review process.

**References**

Bauckhage, C. “Insights into Internet Memes,” in International AAAI Conference on Weblogs and Social Media Insights, 2011.

Bojanowski, P., E. Grave, A. Joulin, and T. Mikolov, “Enriching Word Vectors with Subword Information,” Trans. Assoc. Comput. Linguist., vol. 5, pp. 135–146, Dec. 2017.

Bordogna, G. and G. Pasi, “An approach to identify ememes on the blogosphere,” in Proceedings of the 2012 IEEE/WIC/ACM International Conference on Web Intelligence and Intelligent Agent Technology Workshops, WI-IAT 2012, 2012, pp. 137–141.

Chew and G. Eysenbach, “Pandemics in the age of Twitter: Content analysis of tweets during the 2009 H1N1 outbreak,” PLoS One, vol. 5, no. 11, 2010.

Chagas, V., Freire, F., Ríos, D., & Magalhães, D. (2019). Political memes and the politics of memes: A methodological proposal for content analysis of online political memes. First Monday, 24(2). Retrieved from https://firstmonday.org/ojs/index.php/fm/article/view/7264/7731

Cover, T.M. and P. E. Hart, “Nearest Neighbor Pattern Classification,” IEEE Trans. Inf. Theory, vol. 13, no. 1, pp. 21–27, 1967.

Dalal, N. and B. Triggs, “Histograms of oriented gradients for human detection,” in Proceedings - 2005 IEEE Computer Society Conference on Computer Vision and Pattern Recognition, CVPR 2005, 2005, vol. I, pp. 886–893.

Davison, P. (2012). The language of internet memes. In Mandiberg, M (Ed.) The social media reader, 120-134. New York: NYU Press.

Fang, H. et al., “From captions to visual concepts and back,” Proc. IEEE Comput. Soc. Conf. Comput. Vis. Pattern Recognit., vol. 07-12-June, pp. 1473–1482, 2015.

García, D. (2014). Las imágenes macro y los memes de Internet: posibilidades de estudio desde las teorías de la comunicación. Revista de Tecnología y Sociedad, 4 (6), 1-7.

He, K., X. Zhang, S. Ren, and J. Sun, “Deep residual learning for image recognition,” in Proceedings of the IEEE Computer Society Conference on Computer Vision and Pattern Recognition, 2016, vol. 2016-December, pp. 770–778.

Ferrara, E., M. JafariAsbagh, O. Varol, V. Qazvinian, F. Menczer, and A. Flammini, “Clustering memes in social media,” in Proceedings of the 2013 IEEE/ACM International Conference on Advances in Social Networks Analysis and Mining, ASONAM 2013, 2013, pp. 548–555.

Finkelstein, S. Zannettou, B. Bradlyn, and J. Blackburn, “A Quantitative Approach to Understanding Online Antisemitism,” Sep. 2018.

Gleick, J. The information: a history, a theory, a flood. Pantheon Books, 2011.

Goodfellow, I., Y. Bengio, and A. Courville, Deep Learning. 2016.

Hochreiter and J. Schmidhuber, “Long Short-Term Memory,” Neural Comput., vol. 9, no. 8, pp. 1735–1780, Nov. 1997.

JafariAsbagh, M,. E. Ferrara, O. Varol, F. Menczer, and A. Flammini, “Clustering memes in social media streams,” Soc. Netw. Anal. Min., vol. 4, no. 1, pp. 1–13, Jan. 2014.

Mikolov,T., T. Mikolov, W. Yih, and G. Zweig, “Linguistic regularities in continuous space word representations,” 2013.

Milo, T., A. Somech, and B. Youngmann, “Simmeme: A search engine for internet memes,” in Proceedings - International Conference on Data Engineering, 2019, vol. 2019-April, pp. 974–985.

Miech, A., I. Laptev, and J. Sivic, “Learning a Text-Video Embedding from Incomplete and Heterogeneous Data,” Apr. 2018.

Peirson. A.L. and E. M. Tolunay, “Dank Learning: Generating Memes Using Deep Neural Networks,” Jun. 2018.

Pennington, J, J. Pennington, R. Socher, and C. D. Manning, “Glove: Global vectors for word representation,” IN EMNLP, 2014.

Schroff, F., D. Kalenichenko, and J. Philbin, “FaceNet: A unified embedding for face recognition and clustering,” in Proceedings of the IEEE Computer Society Conference on Computer Vision and Pattern Recognition, 2015, vol. 07-12-June-2015, pp. 815–823.

Quinlan, “Induction of Decision Trees,” Mach. Learn., vol. 1, no. 1, pp. 81–106, 1986.

Schölkopf, B. “SVMs - A practical consequence of learning theory,” IEEE Intell. Syst. Their Appl., vol. 13, no. 4, pp. 18–21, Jul. 1998.

Theodoridis, S., A. Pikrakis, K. Koutroumbas, and D. Cavouras, Introduction to Pattern Recognition: A Matlab Approach. Elsevier Inc., 2010.

Truong, B. Q.,A. Sun, and S. S. Bhowmick, “CASIS: A system for concept-aware social image search,” in WWW’12 - Proceedings of the 21st Annual Conference on World Wide Web Companion, 2012, pp. 425–428.

Tsur, O. and A. Rappoport, “Don’t Let Me Be #Misunderstood: Linguistically Motivated Algorithm for Predicting the Popularity of Textual Memes,” in Proceedings of the Ninth International AAAI Conference on Web and Social Media, 2015.

We Are Social (2019). Global social media research summary 2019. Smart Insights. Retrieved from https://www.smartinsights.com/social-media-marketing/social-media-strategy/new-global-social-media-research/

Yan, F. and K. Mikolajczyk, “Deep correlation for matching images and text,” in Proceedings of the IEEE Computer Society Conference on Computer Vision and Pattern Recognition, 2015, vol. 07-12-June-2015, pp. 3441–3450.

Zannettou, S. et al., “On the origins of memes by means of fringe web communities,” in Proceedings of the ACM SIGCOMM Internet Measurement Conference, IMC, 2018, pp. 188–202.